# The Truth and Nothing but the Truth: Multimodal Analysis for Deception Detection

Mimansa Jaiswal
Institute of Engineering and Technology
DA University
Indore, India
mimansa.jaiswal@gmail.com

Sairam Tabibu
Electronics Engineering
IIT BHU
Varanasi, India
t.sairam.ece13@itbhu.ac.in

Rajiv Bajpai
School of Computer Science and Engineering
Nanyang Technological University
Singapore
rbajpai@ntu.edu.sg

*Abstract*— **We propose a data-driven method for automatic deception detection in real-life trial data using visual and verbal cues. Using OpenFace with facial action unit recognition, we analyze the movement of facial features of the witness when posed with questions and the acoustic patterns using OpenSmile. We then perform a lexical analysis on the spoken words, emphasizing the use of pauses and utterance breaks, feeding that to a Support Vector Machine to test deceit or truth prediction. We then try out a method to incorporate utterance-based fusion of visual and lexical analysis, using string based matching.**

*Keywords— face tracking, support vector machine, multimodal analysis, deception, POS weighted vector*

## I. INTRODUCTION

Trials and verdicts are often a common occurrence in any governance around the world. With more money and fame at stake, however, there has been a great influx of using deceptive statements, often in the form of perjuring themselves or in a bargain. Hence, the need of implementing computational methods that can evaluate the honesty of provided testimonies has arisen. The initial work in this field aims to provide a support system to the judiciary for evaluating the credibility and truthfulness of the testimony given by the witness at stand.

Accurate deception detection helps not only in trials but also in weeding out potential suspects during routine consulate interviews [1]. Currently the most popular means of detecting deception is polygraph or lie-detector machines which monitors heartbeat and physical cues. The system in use faces many problems, especially because it is overt in nature i.e. the subject knows that it is being monitored and can therefore change or control his/her behavior and symptoms either with training or with medication.

Secondly, the polygraph machine requires cooperation from the subject and needs to be used by a trained person, which under some cases can either be corrupted or be inadequate. It relies on precise calibration which may or may not be possible under every judicial circumstance.

In response, researchers have always been trying to automate the process of detecting deceptive statements or deceptive behavior when the subject is under questioning [2]. Such a facility would help them not only detect deception but also find the key-points of notice, such that other people could be trained to look out for it. The most popular way for detecting deception depends on either agitation or over-control [3]. This is explained as: when a person tries to lie his way out, he either forms very controlled sentences or gets nervous and tries to deviate the topic. But the biggest concern in this generalization is the loss of individuality amongst the tested participants.

The biggest drawback in automated deception detection has been the lack of real data for its training purposes. The data in multi-modal scenarios (like affect detection, or emotion detection) has always been that though there are datasets that are acted out under controlled settings, there has been a dearth of data from real situations. Acted out datasets have been said to not accurately represent a factual situation, for example, when a person is angry, they often tend to cry, leading the emotion detected to be confused with sad, while the actors did not take that into account, misleading the model.

The rest of the paper is organized as follows: Section II lists related works in multimodal analysis and emotion detection; Section III presents the proposed method; Section IV provides experimental results; finally, Section V concludes the paper and offers pointers for future work.

## II. RELATED WORK

In this paper, we attempt to build an automatic multimodal deception analysis system, which we believe is one of the few initial attempts. Some of the most prominent works are in the field of multimodal sentiment analysis. Poria [31-33], Pérez-Rosas [34], and Cambria [35, 36] successfully made use of many techniques such as utterance-level frameworks, concept-level engines, ensembles, etc. On the other hand, while there has been work in the use of court trial transcripts, none of them took video or audio into account. The dataset we use is of real-life court trial snippets which has been used to perform multi-modal analysis [4].

In this paper, we aim to analyze the significance of various lexical and visual features that influence a deception model. The presence or absence of a facial feature is manually noted and compared to the human accuracy, which according to the studies is just higher than a random chance. Several other researchers have tried various other ways of detecting deception, namely fMRI [6-8], while others tried to understand behavioral indicators that defined the line between a person speaking truth or a lie [1]. To understand the change in behavior, one needs to establish a baseline and hence study subject specific models. Tsechpenakis et al. [10] extend the work of [9], translating blob features into illustrator and adaptor behaviors and combining these via a hierarchical HMM [11] to decide if the subject falls into the mentioned indicators, i.e., agitated, relaxed or over-controlling.

Reliance on summary functions like mean, median and mode glosses over the abruptness of a subject, which may occur when the lie is stowed between the truth, as used in [9, 10]. Ekman and Friesen call phenomenon as leakage [12]. OpenFace takes into account the transitions between facial features in a video segment which we try to incorporate in our second experiment.

## III. METHOD

We already mentioned in the abstract section, that the method we use aims to automatically detect FAUs and then, use the generated features in concatenation with the lexical features for building a Support Vector Machine model. Next, we outline the dataset we used and the steps that we undertook for analysis.

### A. Dataset Description

The dataset comprised of 61 deceptive and 60 truthful videos sourced from various Youtube channels. As mentioned in [4], three different trial outcomes were used to correctly label a certain trial video clip as deceptive or truthful: guilty verdict, non-guilty verdict, and exoneration. Thus, for guilty verdicts, deceptive clips were collected from a defendant in a trial and truthful videos were collected from witnesses in the same trial. In some cases, deceptive videos are collected from a suspect denying a crime he committed and truthful clips are taken from the same suspect when answering questions concerning some facts that were verified by the police as truthful. Exoneration testimonies are collected as truthful statements.

Clips containing exonerates testimonies are obtained from "The Innocence Project" website. The average length of the videos in the dataset is 28.0 seconds. The average video length is 27.7 seconds and 28.3 seconds for truthful and deceitful videos. The dataset is then annotated for visual features and transcribed to derive linguistic features. All the video clips were transcribed via crowd sourcing using Amazon Mechanical Turk and word repetitions and fillers such as um, ah, and uh, as well as indicate intentional silence using ellipsis were included. The truthful and deceitful statements can be found as sample transcripts in [4]. We did not use all of the videos for some were either too short (the person in question appeared only for a few seconds) or there were too many people and OpenFace was unable to recognize the subject.

We recognize this as a limitation of our study and aim to improve subject recognition, especially in news interviews by the mode of speaking turns. If we would have included these videos, we would be just relying upon a particular modality, which would not cater to the aim of our experiment.

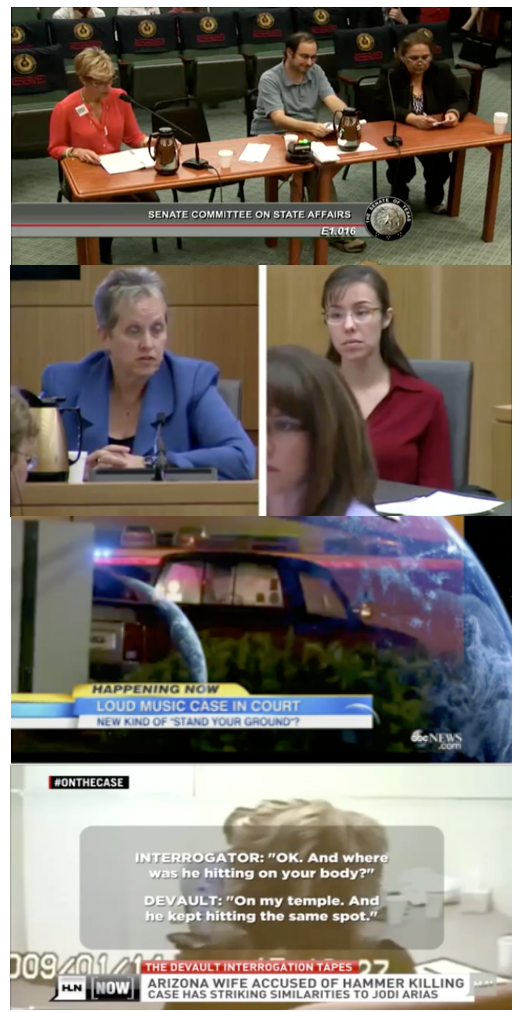

**Fig. 1.** From top bottom, static images from videos that were not used for analysis because (1, 2) too many people to analyze (3, 4) the subject cannot be identified.

We found 8 such videos in deceptive category and 4 of them in truthful category. To equate the number of videos in each category, we brought down the number of videos analyzed to 50 for both of them, by manually going through the Facial Action Units [14] recognized video generated by OpenFace [13].

*B. Visual Feature Extraction*

We then mapped the gestures derived from open face to those annotated in the dataset to find the agreement value. Because the original dataset consisted of manually marked annotations, we consider them to be the baseline for analysis.

The AUs used in OpenFace for this purpose are AU1, AU2, AU4, AU5, AU6, AU7, AU9, AU10, AU12, AU14, AU15, AU17, AU20, AU23, AU25, AU26, AU28 and AU25.

| Annotation | OpenFace |
|---|---|
| Eyebrows | Inner brow raiser (AU1) |
| | Outer brow raiser (AU2) |
| | Brow lowerer (AU4) |
| Eyes | Blink (AU45) |
| | Lid tightener (AU7) |
| Mouth | Lip tightener (AU23) |
| | Lip part (AU25) |
| | Jaw drop (AU26) |
| | Lip suck (AU28) |

**Table 1.** Example mapping between annotation category and OpenFace category

The presence of these AUs in the feature set correctly corresponded to the human-annotated category at an average of 76%. We consider this to be a feasible value because as reported by R. Mihalcea [4], the inter-annotator agreement (human agreement) also averages around 71%.

OpenFace extracted frame-wise features, which we then collated into one feature representing a single video using threshold presence for each AU. We tried different thresholds for the ones that were defined by intensity, but finally settled upon 3.0 as a good measure. This generated a visual vector modal for analysis of 18 dimensions.

*C. Lexical Feature Extraction*

For verbal feature extraction, bag of words model was used. We initially build a vocabulary set consisting of unigrams and their frequency was mapped. The vocabulary set finally did not contain articles, prepositions or filler words.

We specially did not remove the repetitions and pauses, for we wanted to understand their effect in determination of truth and lies. All the words that had a frequency below 5 were removed from the set.

The remaining words represent the unigram features, which are then associated with a value corresponding to the frequency of the unigram inside each utterance transcription. These simple weighted unigram features have previously been shown to achieve state of art performance using Support Vector Machines (SVMs) [15].

To improve the above vector set, we added emotional information from SenticNet [37] and its extensions [38, 39], a concept-level knowledge base for sentiment analysis that provides both semantic and affective information associated to words and multiword expressions by means of commonsense computing [40, 41], affective reasoning [42, 43], and sentic computing [44, 45]. We also added weightage to POS, increasing the weight by a factor of 1.4 for pronouns and 1.2 for adjectives. The idea was to make the model context-neutral, because many of the words like "sofa", "men", "brother", did not significantly add to deception detection. Finally, we increased the weight of pauses by 1.6, which can be an important factor in nervous or over-controlled situations [29].

Other than weighted unigram features, we tried to make use of word embedding for the same purpose to generate a final vector, but the change did not reflect any significant results. We also tried to use convolutional neural networks and recurrent neural networks to develop the vector model, but probably due to insufficient number of words in vocabulary and the short length of documents, it did not yield a significant positive change.

*D. Acoustic Feature Extraction*

Audio based features were extracted using openSMILE [28] to extract basic features like Mel-frequency coefficient, harmonics-to-noise ratio, jitter (jitter has been proven to be a good indicator of nervousness, which is one of the categories of lying as stated by Zhang Z. [1]). The feature vector comprised of 28 dimensions finally, when used. We include prosody, energy, voicing probabilities, spectrum, and cepstral features.

- Prosody features. These include intensity, loudness, and pitch that describe the speech signal in terms of amplitude and frequency.

- Energy features. These features describe the human loudness perception.

- Voice probabilities. These are probabilities that represent an estimate of the percentage of voiced and unvoiced energy in the speech.

- Spectral features. The spectral features are based on the characteristics of the human ear, which uses a nonlinear frequency unit to simulate the human auditory system. These features describe the speech formants, which model spoken content and represent speaker characteristics.

- Cepstral features. These features emphasize changes or periodicity in the spectrum features measured by frequencies; we model them using 12 Mel-frequency cepstral coefficients that are calculated based on the Fourier transform of a speech frame.

## IV. EXPERIMENTS

We then performed individual modality accuracy analysis and fusion analysis. We tested three types of fusion, namely, early fusion, decision level fusion and utterance based fusion. When we performed individual modality analysis, the acoustic part performed the worst whereas the visual part performed the best. We think that the acoustic part gives low accuracy because (a) Most of the videos accounted for more than one person speaking (especially because of interviewer and interviewee) (b) The recordings were mostly from courtroom, which inherently had a lot of noise due to public proceedings and the inflections in voice weren't that useful.

From each transcript, we extract the linguistic, acoustic, and visual features described above, which are then combined using the early fusion (or feature-level fusion) approach [30]. In this approach, the features collected from all the multimodal streams are combined into a single feature vector. We also tried decision level fusion. In decision-level fusion, we obtained feature vectors from the above-mentioned methods but did concatenate the vector and rather used separate classifiers. This is also called as late-fusion.

The output of each classifier was treated as a classification score, and we obtained 2 probability score from each classifier and used argmax summation over weight*score which led us to the final decision. Thirdly, we tried building an utterance level classifier, but did not want to switch to manual mode as tried by [27].

Therefore, we distributed the frames obtained using OpenFace as a proportion of the ratio of word length to the document length and then performed feature extraction over each utterance using similar method. We then used feature level fusion to combine modalities of each utterance and build an SVM classifier.

| Modality | Accuracy |
|---|---|
| Baseline (Human Annotators) | 55.93% |
| Baseline (Manually extracted gestures + L) | 75.20% |
| One modality at a time | |
| Lexical | 66.12% |
| Acoustic | 34.23% |
| Visual | 67.20% |
| V+L+A | |
| Feature-level fusion | **78.95%** |
| Decision level fusion | 76.12% |
| Utterance based feature fusion | 74.02% |

**Table 2.** Results of unimodal and multimodal models using different fusion techniques

The average human judgement accuracy ranged from 53% to 60% [4]. In our case, the method of feature-level of fusion as automatic system works best at an average of 78.95%. The accuracy in truth videos was found to be higher at 81.10% while that in deceptive videos was lower at around 76.80%.

The baseline model [4] possesses an accuracy of 75.20% over full video while it has 60.33% accuracy single text modality and 68.59% accuracy over silent video (visual modality). In our experiments, though we receive a higher accuracy over text (66.12%), it is lower in visual modality by a non-significant amount (67.20%). The probable reason could be that in the baseline model, the visual cues are extracted manually, while we extract them automatically, which eventually decreases the accuracy. We could not use the gesture annotations because, then, the system would not be fully automatic.

## V. CONCLUSION

We have proposed a method for automatic deception detection using three modalities: visual, acoustic and lexical. In particular, we have automated the gesture recognition and mapping and improved the textual analysis using weighted features and POS highlighting.

In the future, we plan to take a concept-level approach [46] to the detection of deception for better integration with SenticNet, which contains multiword expressions in stead of affect words, and include the use of linguistic patterns [47, 48] to improve the detection accuracy. Additionally, we plan to integrate in our framework modules for personality recognition [49] and multimodal emotion recognition [50].

Gaze has been said to be an important indicator of truthfulness or an indication of human credibility. We did not incorporate gaze into our experiments though OpenFace provides a medium for tracking it, and would like to do that in future. Secondly, the method of fusion for utterances is based on averages rather than actual words. This discounts the duration of pauses and the effect of pause duration on classifier. Thirdly, the acoustic part needs to take into account (a) noise reduction (b) lawyer/witness switching which we aim to incorporate into our future experiments.